\documentclass[runningheads]{llncs}
\usepackage{graphicx}

\usepackage{tikz}
\usepackage{comment}
\usepackage{amsmath,amssymb} %
\usepackage{color}
\usepackage{letltxmacro}
\usepackage[width=122mm,left=12mm,paperwidth=146mm,height=193mm,top=12mm,paperheight=217mm]{geometry}
\usepackage{cleveref}
\usepackage{booktabs}
\usepackage[T1]{fontenc}
\usepackage{setspace}
\begin{document}
\pagestyle{headings}
\mainmatter

\title{DDNeRF: Depth Distribution\\ Neural Radiance Fields} %

\titlerunning{DDNeRF: Depth Distribution Neural Radiance Fields} 
\authorrunning{David Dadon, Ohad Fried, Yacov Hel-Or} 

\author{David Dadon, Ohad Fried, Yacov Hel-Or\\
School of Computer Science, Reichman University, Herzliya, Israel \\
\textit {david.dadon@post.idc.ac.il , ofried@idc.ac.il, toky@idc.ac.il }}  
\institute{}

\maketitle

\newcommand{\betweencellpdf}{\ensuremath{h^c}}
\newcommand{\betweencellpdffine}{\ensuremath{h^f}}
\newcommand{\approxbetweencellpdffine}{\ensuremath{\hat{h}^f}}
\newcommand{\incellpdf}{\ensuremath{f}}
\newcommand{\incellpdfnormalized}{\ensuremath{f'}}

\newcommand{\incellcdf}{\ensuremath{F}}

\newcommand{\totalpdf}{\ensuremath{f_{dd}}}
\newcommand{\totalcdf}{\ensuremath{F_{dd}}}

\newcommand{\ignorethis}[1]{}
\newcommand{\redund}[1]{#1}

\newcommand{\apriori    }     {\textit{a~priori}}
\newcommand{\aposteriori}     {\textit{a~posteriori}}
\newcommand{\perse      }     {\textit{per~se}}
\newcommand{\naive      }     {{na\"{\i}ve}}
\newcommand{\Naive      }     {{Na\"{\i}ve}}
\newcommand{\Identity   }     {\mat{I}}
\newcommand{\Zero       }     {\mathbf{0}}
\newcommand{\Reals      }     {{\textrm{I\kern-0.18em R}}}
\newcommand{\isdefined  }     {\mbox{\hspace{0.5ex}:=\hspace{0.5ex}}}
\newcommand{\texthalf   }     {\ensuremath{\textstyle\frac{1}{2}}}
\newcommand{\half       }     {\ensuremath{\frac{1}{2}}}
\newcommand{\third      }     {\ensuremath{\frac{1}{3}}}
\newcommand{\fourth     }     {\ensuremath{\frac{1}{4}}}

\newcommand{\Lone} {\ensuremath{L_1}}
\newcommand{\Ltwo} {\ensuremath{L_2}}

\newcommand{\degree} {\ensuremath{^{\circ}}}

\newcommand{\mat        } [1] {{\text{\boldmath $\mathbit{#1}$}}}
\newcommand{\Approx     } [1] {\widetilde{#1}}
\newcommand{\change     } [1] {\mbox{{\footnotesize $\Delta$} \kern-3pt}#1}

\newcommand{\Order      } [1] {O(#1)}
\newcommand{\set        } [1] {{\lbrace #1 \rbrace}}
\newcommand{\floor      } [1] {{\lfloor #1 \rfloor}}
\newcommand{\ceil       } [1] {{\lceil  #1 \rceil }}
\newcommand{\inverse    } [1] {{#1}^{-1}}
\newcommand{\transpose  } [1] {{#1}^\mathrm{T}}
\newcommand{\invtransp  } [1] {{#1}^{-\mathrm{T}}}
\newcommand{\relu       } [1] {{\lbrack #1 \rbrack_+}}

\newcommand{\abs        } [1] {{| #1 |}}
\newcommand{\Abs        } [1] {{\left| #1 \right|}}
\newcommand{\norm       } [1] {{\| #1 \|}}
\newcommand{\Norm       } [1] {{\left\| #1 \right\|}}
\newcommand{\pnorm      } [2] {\norm{#1}_{#2}}
\newcommand{\Pnorm      } [2] {\Norm{#1}_{#2}}
\newcommand{\inner      } [2] {{\langle {#1} \, | \, {#2} \rangle}}
\newcommand{\Inner      } [2] {{\left\langle \begin{array}{@{}c|c@{}}
                               \displaystyle {#1} & \displaystyle {#2}
                               \end{array} \right\rangle}}

\newcommand{\twopartdef}[4]
{
  \left\{
  \begin{array}{ll}
    #1 & \mbox{if } #2 \\
    #3 & \mbox{if } #4
  \end{array}
  \right.
}

\newcommand{\fourpartdef}[8]
{
  \left\{
  \begin{array}{ll}
    #1 & \mbox{if } #2 \\
    #3 & \mbox{if } #4 \\
    #5 & \mbox{if } #6 \\
    #7 & \mbox{if } #8
  \end{array}
  \right.
}

\newcommand{\len}[1]{\text{len}(#1)}

\newlength{\w}
\newlength{\h}
\newlength{\x}

\definecolor{darkred}{rgb}{0.7,0.1,0.1}
\definecolor{darkgreen}{rgb}{0.1,0.6,0.1}
\definecolor{cyan}{rgb}{0.7,0.0,0.7}
\definecolor{otherblue}{rgb}{0.1,0.4,0.8}
\definecolor{maroon}{rgb}{0.76,.13,.28}
\definecolor{burntorange}{rgb}{0.81,.33,0}

\ifdefined\ShowNotes
  \newcommand{\colornote}[3]{{\color{#1}\textbf{#2} #3\normalfont}}
\else
  \newcommand{\colornote}[3]{}
\fi

\newcommand {\todo}[1]{\colornote{cyan}{TODO}{#1}}
\newcommand {\ohad}[1]{\colornote{otherblue}{OF:}{#1}}
\newcommand {\toky}[1]{\colornote{darkgreen}{YH:}{#1}}
\newcommand {\dudi}[1]{\colornote{burntorange}{DD:}{#1}}

\newcommand {\reqs}[1]{\colornote{red}{\tiny #1}}

\newcommand {\new}[1]{\colornote{red}{#1}}

\newcommand*\rot[1]{\rotatebox{90}{#1}}

\newcommand {\newstuff}[1]{#1}

\newcommand\todosilent[1]{}

\newcommand{\woBGmask}{{w/o~bg~\&~mask}}
\newcommand{\woMask}{{w/o~mask}}

\providecommand{\keywords}[1]
{
  \textbf{\textit{Keywords---}} #1
}

\newcommand {\shortcite}[1]{\cite{#1}}

\begin{abstract}
In recent years, the field of implicit neural representation has progressed significantly. Models such as neural radiance fields (NeRF) \cite{mildenhall2020nerf}, which uses relatively small neural networks, can represent high-quality scenes and achieve state-of-the-art results for novel view synthesis. Training these types of networks, however, is still computationally very expensive. We present depth distribution neural radiance field (DDNeRF), a new method that significantly increases sampling efficiency along rays during training while achieving superior results for a given sampling budget. DDNeRF achieves this by learning a more accurate representation of the density distribution along rays. More specifically, we train a coarse model to predict the internal distribution of the transparency of an input volume in addition to the volume's total density. This finer distribution then guides the sampling procedure of the fine model. This method allows us to use fewer samples during training while reducing computational resources.   

\end{abstract}

\section{Introduction}

\begin{figure}
\centering
\includegraphics[width=0.9\textwidth]{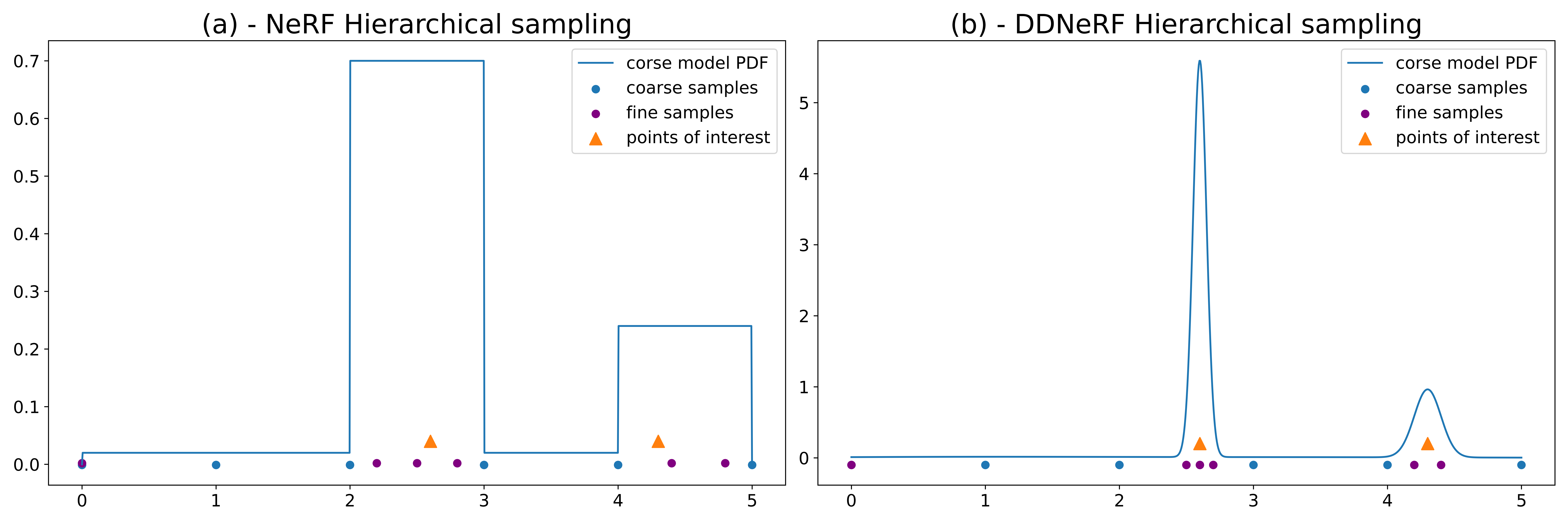}
\caption{\textbf{Hierarchical sampling}: The horizontal axis represents the depth axis of the ray for a scene with max depth of 5. First, the coarse samples (blue dots) are taken; then the density values are transformed into a PDF function (blue line). The fine model samples (purple dots) are taken with respect to the coarse model PDF. The orange rectangle represents points that have an influence on the pixel color (similar in both plots). The left plot illustrates the sampling procedure in the regular model, The right plot illustrates our scheme. Notice how when using our representation, the finer samples are concentrated around the informative areas.}
\label{fig:hierarchical_sampling}
\end{figure}

The field of implicit representation for 3D objects and scenes has been growing rapidly in the last several years. Methods such as Occupancy Networks \cite{Occupancy_Networks} and DeepSDF \cite{park2019deepsdf} (Signed Distance Function) have achieved state-of-the-art results in 3D reconstruction, which led to increased interest in this field. The two main advantages of implicit representation are compactness and continuity (compared to explicit representation methods such as meshes or voxels that are discrete and less compact). It also enables us the 3D shape of the represented object for every level of detail (LOD) to be extracted by increasing/decreasing the number of samples in space. Due to their performance and accuracy, implicit methods became very popular, adopted by many papers and various domains.\\
Neural Radiance Fields (NeRF) \cite{mildenhall2020nerf} use the same architecture as DeepSDF to represent a scene as a radiance field by answering the following query: given an $(x,y,z)$ location and a viewing direction $(\phi, \theta)$, what is the RGB color and the density $\sigma$ 
in this location? When rendering an image, a pixel color is evaluated by sampling points along the ray from the center of projection (COP) that passes through the pixel and applying a ray marching rendering technique for volume rendering \cite{10.1145/74333.74359}. \\ 
At the time this method was published it achieved cutting-edge results for novel view synthesis. This led to what is called the ``NeRF explosion''. In the past two years, numerous follow-up works improved the NeRF model and extended it to new domains. We review a few of those works briefly in the Section 2. \\
Nevertheless, NeRF has one major drawback: its training time and space requirements. Because the quality of the model depends on the number of samples drawn along each ray (more samples produce better models), the training process has a trade-off between efficiency (number of samples) and output quality.\\
Most NeRF models use two-stage hierarchical sampling techniques. The first stage (coarse model) samples uniformly with respect to the depth axis along the ray and divides the ray into intervals according to these samples. The opacity ($\alpha$) and the total transparency of each interval $i$ are used to determine the amount of influence $w_i$ of each interval on the pixel color (see \cref{weight_calc}). The $w_i$'s values are normalized and interpreted as a piecewise-constant PDF (or discrete PDF) between the intervals. The second stage (fine model) samples points according to this PDF function.  \Cref{fig:hierarchical_sampling} (a) illustrates this method.\\
In this paper we propose to represent the PDF of the first hierarchical sampling stage as a combination of Gaussian distributions and we will show its advantages over the piecewise-constant PDF. The input of the model is a specific sub-section of the ray (interval), and the output is the Gaussian distribution parameters ($\mu,\sigma$) of the density influence in that interval \textendash$~$in addition to the color and the total density. By using this technique, we achieve a more accurate density representation along the rays, which will allow us to receive more accurate samples in the second stage of the hierarchical sampling.  \Cref{fig:hierarchical_sampling} (b) illustrates our method. We will demonstrate and analyze its superiority over the piecewise-constant PDF representation for a variety of domains and sampling budgets. Our model and PDF representation are versatile and can be applied to almost each of the existing NeRF Models. \textbf{Our main contributions} can be described as follow:
\begin{enumerate}
   \item A finer and more continuous representation of the density distribution along the ray in NeRF based models, which leads to better results for a given number of samples. This allows us to train the model with less computational resources.
    \item A novel distribution estimation (DE) loss, which provides an additional path for information to flow from the fine to the coarse model and improve the overall model performance.
\end{enumerate}

\section{Related Works}

\subsubsection{Implicit 3D representation:}
Two of the first methods to achieve very good results for implicit representation of 3D objects were Occupancy Networks \cite{Occupancy_Networks}, which, for a 3D point input, was trained to answer the query: ``is this point inside or outside the 3D object?'' and DeepSDF \cite{park2019deepsdf}, which was trained to answer the query: given an $(x,y,z)$ location in space, what is the distance to the zero-level surface?, where positive and negative distances represent whether the point is located outside or inside the shape. By answering the above queries for enough 3D points in a space, combined with a variant of the marching cube algorithm, the 3D shape of an object can be extracted. These methods became very popular due to their good results, compactness and continuity characteristics. Additional papers, such as Pifu \cite{saito_pifu:_2015} followed and these too tried to answer similar queries to extract 3D shapes. More advanced implicit models, e.g., SAL \cite{Atzmon_2020_CVPR} and SALD \cite{atzmon2021sald}, were developed to train directly from the row 3D data without using ground truth (GT).
 
\subsubsection{NeRF models:}
As described above, the NeRF \cite{mildenhall2020nerf} method uses a neural network to imply implicit representation of radiance field for volumetric rendering. It gets an $(x,y,z)$ location and a view direction $(\phi, \theta)$, and predicts the RGB color and the density $\alpha$ in this location. When this method was published, it achieved state-of-the-art results in the task of novel view synthesis. In the past two years, many works extended the NeRF model to additional tasks and domains. NeRF++ \cite{kaizhang2020} extends the model to unbounded-real world scenes using additional neural network for background modeling and new background parametrization. \cite{martinbrualla2020nerfw} extend the model for unconstrained image collection and \cite{DBLP:journals/corr/abs-2011-13961} extend it for dynamic scenes. MipNeRF \cite{barron2021mipnerf} addresses the model aliasing problem with different resolution images. Many works also tried to decrease the computational resources required during training and, especially, the amount of inference time demanded \cite{neff2021donerf} \cite{kangle2021dsnerf} \cite{DBLP:journals/corr/abs-2103-13744} \cite{liu2020neural}.\\ 
The connection between sampling around informative depth locations and computational complexity appeared in some of the above-mentioned works. DSNeRF \cite{kangle2021dsnerf} uses some prior depth information to improve training time and output quality when training with a small number of images. NSVF \cite{liu2020neural} uses sparse voxel fields to achieve better sampling locations. DONeRF \cite{neff2021donerf} improves inference time by using a depth oracle for sampling in informative locations. The depth oracle is trained with GT Depth (or a trained NeRF model) to predict an accurate location for the second stage sampling. DONeRF \cite{neff2021donerf} also uses log-sampling and space warping techniques to increase model quality on areas far from the camera.

\section{Problem Definition}

\begin{figure}
\centering
\includegraphics[width=0.9\textwidth]{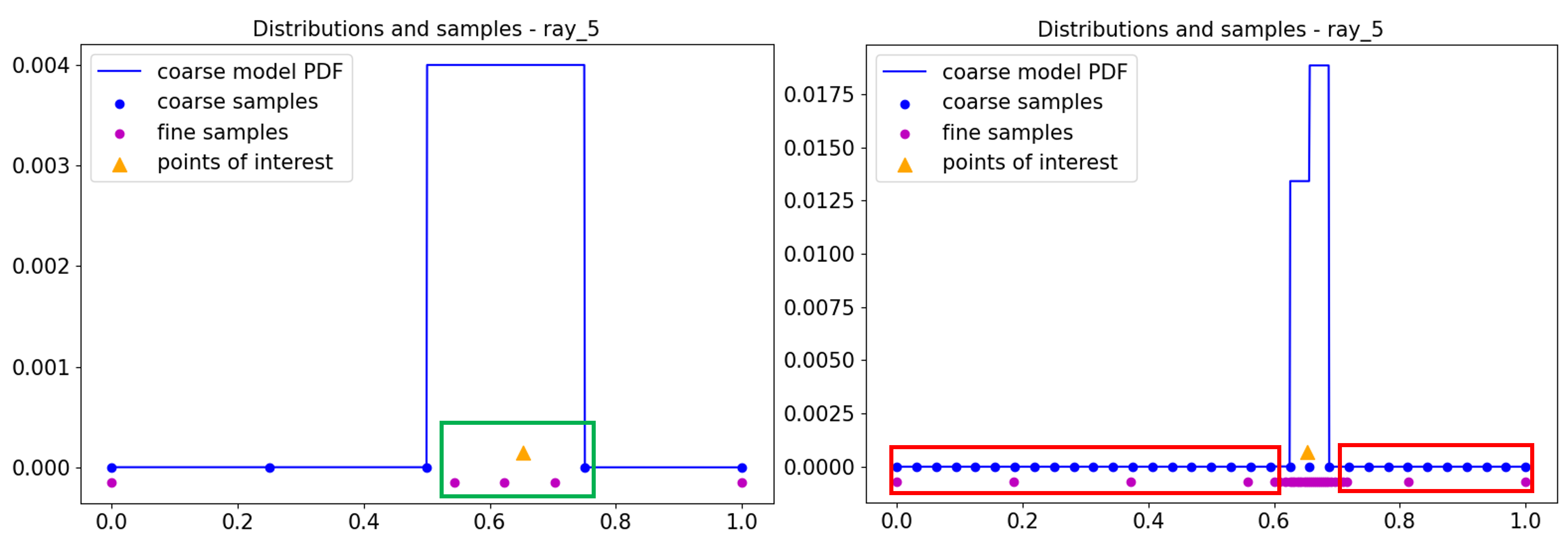}
\caption{Left plot: Limited depth resolution when using a small number of samples (inside the green rectangle). Right plot: Many samples with zero contribution (inside the red rectangles).}
\label{fig:problem}
\end{figure}

When looking deeper into the NeRF hierarchical sampling strategy we observed two inherent disadvantages. The first one is that for n samples, the second pass sampling resolution cannot be better than $\frac{1}{n^2}$ of the scene depth. In other words, even if the first pass predicts that 100 percent of the samples of the second pass should be placed in a single interval, because the PDF along the ray is discrete, the finer sampling will sample this interval uniformly. \\
To overcome this problem we are forced to use a large number of samples during training (a small number of samples will lead to a non-accurate depth estimation). Another derivative of this problem is that for a deep or unbounded scene, even when using a large number of samples, the model still struggles to achieve good results and there is a trade-off between background to foreground quality (as shown in Nerf++\cite{kaizhang2020}) as a function of the sample's depth range.\\
The second disadvantage we observed in the traditional NeRF sampling strategy is that most of the samples in the first pass contribute almost nothing to the training process because they predict zero influence from a very early stage in the training until its end. Despite this we still need to use them because of the first problem we mentioned. \Cref{fig:problem} illustrates the inherent trade-off between the two problems.

\section{Our Model}

\subsection{Preliminaries}
Our model is a direct extension of NeRF \cite{mildenhall2020nerf} and MipNeRF \cite{barron2021mipnerf}. For this reason we will start by describing those models in more detail.  

\subsubsection{NeRF}
As mentioned above, NeRF receives a 5D input: $(x,y,z,\theta,\phi)$, and produces a 4D output: $(R, G, B, \sigma)$, where $\sigma$ is the density of the input point that translates later into opacity $\alpha$ (value between 0 and 1) by considering the distance $\delta$ along the ray that is affected by that $\sigma$. So, for sample $i$:\\
\begin{equation}\label{alpha_calc}
\alpha_{i} = 1-\exp(-\sigma_i\delta_i)\; \\
\end{equation}
To avoid confusion with the $\sigma$ in our model that represents standard deviation, we will omit $\sigma$ in the notation from now on and refer directly to $\alpha$. The influence of each sample $i$ on the final color prediction $w_i$, is a combination of the total transparency from sample $i$ to the pixel and sample $i$ opacity value ($\alpha_i$):

\begin{equation}\label{weight_calc}
w_{i} = \alpha_{i}\cdot\displaystyle\prod^{i-1}_{j=0}(1-\alpha_{j})\; \\
\end{equation}

The NeRF architecture is composed of two identical networks (coarse and fine) with eight fully connected (FC) layers. The input is first encoded using positional encoding (PE) and then inserted into the network. As described in section 1 (Introduction), the model use two-stage hierarchical sampling where the $w_i$'s values of the coarse model are normalized and can be interpreted as a discrete PDF $\betweencellpdf$:

\begin{equation}\label{pdf_calc}
\betweencellpdf[i] = \frac{w_i}{\sum^n_{j=0}w_j} 
\end{equation}

The fine model samples the second stage of the hierarchical sampling with respect to $\betweencellpdf$. \Cref{fig:hierarchical_sampling} (a) illustrates this process.
Color rendering is performed using ray marching \cite{10.1145/74333.74359} for volumetric rendering and calculated as follow:

\begin{equation}
\hat{C}(\textbf{r}) = \displaystyle\sum^{n}_{i=1} w_{i}c_{i}\\    
\end{equation}
where $\hat{C}(\textbf{r})$ is the predicted pixel color for ray $\textbf{r}$, $c_{i}$ is the RGB prediction for sample $i$ and $w_{i}$ is the influence that sample $i$ has on the final RGB image. $\hat{C}_c(\textbf{r})$ and $\hat{C}_f(\textbf{r})$ are the coarse and fine model color predictions. The calculations of $w_i$ and $c_i$ are made separately for the coarse and the fine models. The loss function is defined as:
\begin{equation}\label{nerf_loss}
L_{nerf} = \displaystyle\sum_{r\in R} [||C(\textbf{r}) - \hat{C}_f(\textbf{r})||^2  + ||C(\textbf{r}) - \hat{C}_c(\textbf{r})||^2]    
\end{equation}
where R is the rays batch for loss calculation and $C(\textbf{r})$ is the ground truth color. \\
\subsubsection{MipNeRF}
MipNeRF \cite{barron2021mipnerf} is an extension of the regular NeRF model that was suggested to handle aliasing caused when rendering images at different resolutions or in different distances than the images used in the training process. Instead of a line, MipNeRF  refers to a ray as a cone \cite{10.1145/964965.808589} with a vertex in the COP that passes through the relevant pixel with a radius related to the pixel size. The cone is divided into intervals (parts of the cone) along the depth axis and the network receives the encoding of an interval as input. Each ray is divided into $n$ intervals bounded by $n+1$ partitions \{$t_i$\} where interval $i$ is the cone volume bounded between partitions $t_{i}$ and $t_{i+1}$; see \Cref{fig:our model} (1). The bounded volumes are encoded using the novel integrated positional encoding (IPE) method for volume encoding before being passed through the network. With this new ray representation the model can understand the entire volume that will affect the pixel value. MipNeRF uses a single neural network for both the coarse and the fine passes. The rest of the process is similar to the original NeRF. \\
In our model we also use the idea that the model predicts information regarding the interval of a cone and not for a point on a ray. 

\subsection{General Description}  
\begin{figure}
\centering
\includegraphics[width=0.9\textwidth]{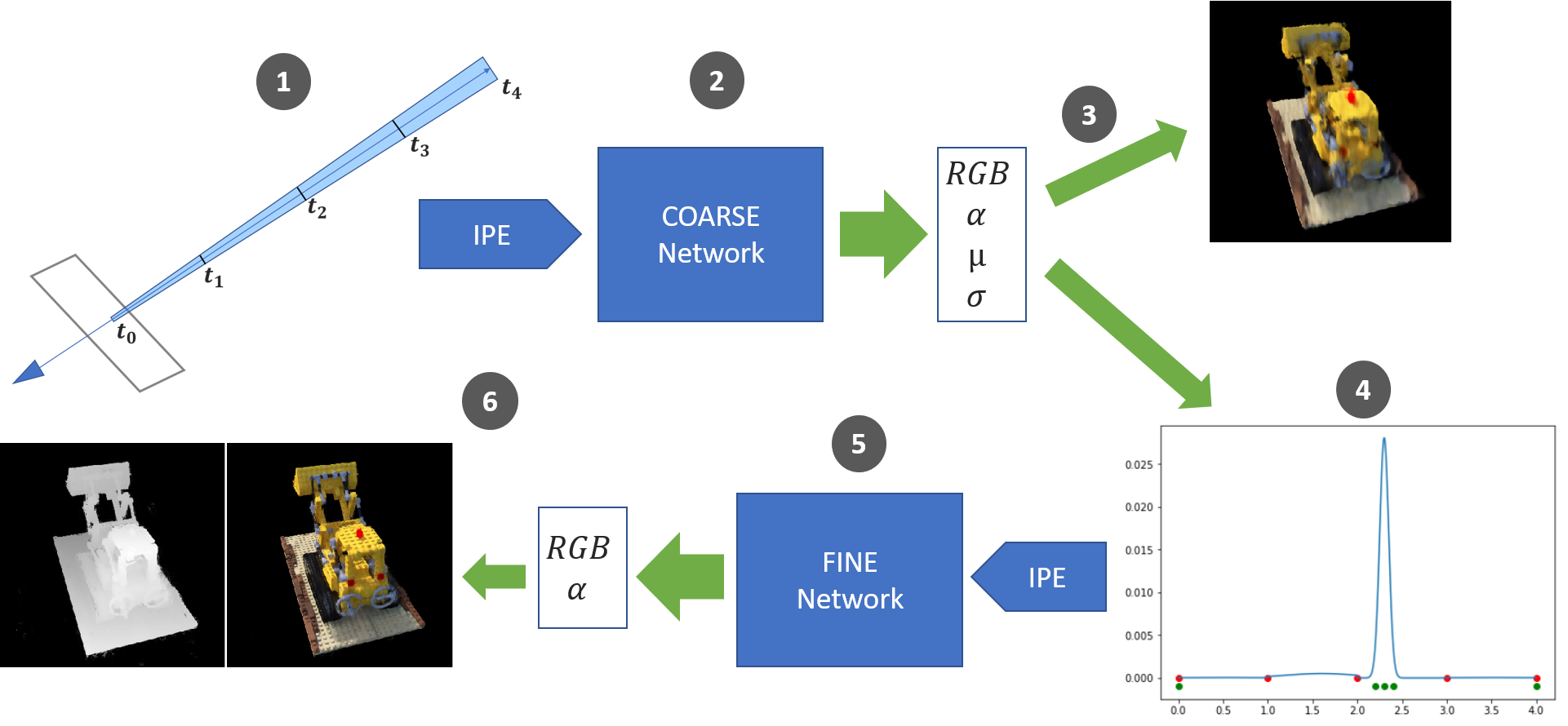}
\caption{\textbf{DDNeRF full pipeline}: (1) Drawing a cone in space and splitting it into relatively uniform intervals along the depth axis. (2) Pass these intervals through an IPE and then through the coarse network to get predictions. (3) Render the coarse RGB image. (4) Approximate the density distribution and include the interval's internal distribution inside the coarse sections boundaries (red dots); then sample the fine samples (green dots). (5) Pass these samples through an IPE and thereafter through the fine network to get predictions. (6) Render the final RGB image and depth map.}
\label{fig:our model}
\end{figure}

As mentioned above, we are trying to extract additional information about the distribution of the density along the ray from our coarse model. We will show that a more accurate estimation of the influence distribution of the density along the ray predicted by the coarse network will lead to better fine samples and improved results.\\
To distinguish between the coarse and fine samples, we denote $T^c=\{t^c_i\}^n_{i=0}$ as the coarse model samples and $T^f=\{t^f_i\}^n_{i=0}$ as the fine model samples.
Similar to MipNeRF, our coarse model gets, as an input, an interval of a cone, but in addition to the regular $RGB\alpha$ output, it also predicts an estimation of the density influence distribution inside this section. More specifically, it predicts the mean $\mu$ and s.t.d. $\sigma$ of the distribution inside that interval. We assume the distribution inside each interval is Gaussian and it does not affect or is affected by adjacent intervals. The importance of this assumption will be explained later in this section.\\
The coarse network learns to predict the distribution by trying to mimic the fine network distribution. We assume that the fine network has a better estimation of the density along the ray. This process will be described in more detail in Section 4.3. The entire pipeline of our model is shown in \Cref{fig:our model}. 
\subsubsection{Architecture:}
We use the MipNeRF\cite{barron2021mipnerf} architecture with two modifications:
(1) We use two different networks for the coarse and fine models - similar to what was done in the original NeRF paper. (2) We change the final FC layer of the coarse model, adding $\mu$ and $\sigma$ to the predictions.

Similar to MipNeRF, we use IPE to encode the input sections before inserting them into the network. Although we chose to use the MipNeRF model, our model can be integrated with any variant of NeRF that uses hierarchical sampling by changing the coarse model.

\subsection{Estimation of Density Distribution}
The predicted $\mu$ and $\sigma$ are limited to be in the range between 0 and 1. Those constraint are implemented by passing the predicted values through the sigmoid activation function. Those values are interpreted relatively to length of the interval. the notation $\mu^r_{i}$, $\sigma^r_{i}$ stands for the relative mean and s.t.d. of interval $i$. The transformation from the relative interpretation to the absolute location and scale along the ray ($\mu_{i}$, $\sigma_{i}$) is calculated as follows:
\begin{equation}
    \mu_{i} = t^c_{i} + \mu^r_{i}\cdot(t^c_{i+1}-t^c_{i})
\end{equation}
\begin{equation}
    \sigma_{i} = \sigma^r_{i}\cdot(t^c_{i+1}-t^c_{i})
\end{equation}

These additional outputs allow us to achieve a finer distribution estimation along the ray. That means that, in addition to the discrete PDF estimation $\betweencellpdf$ between the intervals, we also estimate the distribution inside each interval. The total distribution along the ray is approximated as a combination of Gaussian distributions (one Gaussian for each interval) that allows us to focus the fine samples in a smaller area along the ray. \\
The PDF inside interval $i$ is denoted as $\incellpdf_{i}(t)= \mathcal{N}(t|\mu_{i}, \sigma_{i})$ and its CDF denoted as $\incellcdf_i(t)=\int^{t}_{-\infty}\incellpdf_i(\tau)d\tau$. Because we want $\incellpdf_{i}$ to be bounded inside the interval, we need to truncate the Gaussian to be inside the interval boundaries, normalize it and define a truncated Gaussian distribution \incellpdfnormalized. The truncated Gaussian distribution $\incellpdfnormalized_i$ inside interval $i$ is defined as follows:

\begin{equation}\label{beta_calc}
\incellpdfnormalized_i(t)=\frac{1}{k_i}\cdot\incellpdf_i(t) ~~;~~ k_{i} = F_{i}(t_{i+1})-F_{i}(t_{i})
\end{equation}

\begin{figure}
\centering
\includegraphics[width=1.0\textwidth]{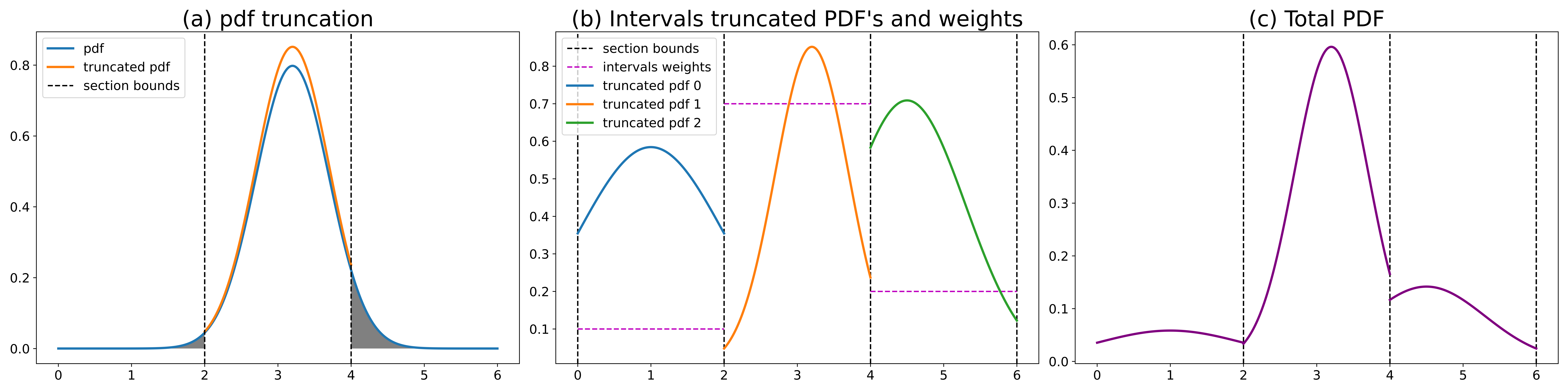}
\caption{(a) The PDF truncation process. The tails that exceed the section boundaries are gray. The blue and orange curves are the PDF before ($f_{i}$) and after ($\incellpdfnormalized_{i}$), i.e., before and after the truncation. (b) Different interval truncated PDF distributions and weights. Each color represents an interval. The vertical dashed lines are the interval bounds and the horizontal lines are the intervals weights (\betweencellpdf). (c) The union of all distributions into one finer distribution \totalpdf.}
\label{fig:total_pdf}
\end{figure}

The truncation procedure is illustrated in \Cref{fig:total_pdf} (a). The discrete function $\betweencellpdf$ between the intervals is calculated as is done in the regular NeRF model (\cref{pdf_calc}).
The total density distribution estimated by the coarse model is a mixture of the truncated Gaussian models when $\betweencellpdf$ is used as the Gaussian weights. This distribution is denoted as \totalpdf. The calculation of $\totalpdf$ is defined in \cref{total_pdf}. The entire procedure illustrated in \Cref{fig:total_pdf}.

\begin{equation}\label{total_pdf}
\totalpdf(t) = \betweencellpdf[i]\cdot \incellpdfnormalized_{i}(t)  ~~~\mbox{when}~~ t \in [t_i, t_{i+1}]
\end{equation}

The main reason we prefer the mixture of truncated Gaussians over the regular mixture of Gaussians is that we want each Gaussian to affect only a single interval. This property is necessary for two reasons. First, because the model calculates each interval independently; we do not want the results of one interval to affect others. Second, assigning each Gaussian to one specific interval allows us to calculate the second pass samples and the additional loss component (described below) efficiently and without requiring significant extra time or memory.   

\subsubsection{Distribution Estimation (DE) Loss:}
To help the coarse network learn to approximate the density distribution, we assume that the density distribution of the fine network is always closer to the real density distribution. Hence we are forcing the predicted coarse distribution to be close to the fine one.\\ 
The fine PDF function $\betweencellpdffine$ is a discrete function computed similarly to \cref{pdf_calc} with respect to density output $\alpha$ of the fine samples $T^f$. We want to estimate $\betweencellpdffine$ by using the coarse model PDF function $\totalpdf$. We use $\approxbetweencellpdffine$ to denote the estimated $\betweencellpdffine$.   
Because $\totalpdf$ is defined for every location on the ray,  we can estimate $\approxbetweencellpdffine$ using $\totalpdf$ and its CDF function $\totalcdf$ as follows:\\
\begin{equation}
\begin{aligned}
     \approxbetweencellpdffine[i] = {} & Pr(t'_{i}\leq t \leq t'_{i+1}) = \int^{t'_{i+1}}_{t'_i}\totalpdf(t)\,dt\ = \totalcdf(t'_{i+1}) - \totalcdf(t'_{i})
\end{aligned}
\end{equation}

We use KL divergence to measure the divergence between the two discrete probabilities  $\betweencellpdffine$ and $\approxbetweencellpdffine$.
Using the KL loss naively tends to push $\mu$ and $\sigma$ toward values close to 0 or 1 and impairs the model convergence (by over-shrinking the Gaussians or leading the model predictions to the vanishing gradient area of the sigmoid function). To avoid these issues, we add two regularization terms encouraging the Gaussian (before truncation) to remain in the center of the interval, with s.t.d. large enough to avoid over-shrinking. This regularization also keeps the model inside the effective range of the sigmoid function. The regularization components of the loss function are $\displaystyle\sum_{i}\mu_{raw}^2$ and $\displaystyle\sum_{i}\sigma_{raw}^2$ where $\mu_{raw}$ and $\sigma_{raw}$ are the model outputs values before passing through the sigmoid function to limit the range to be between 0 to 1. The overall DE loss is defined as:

\begin{equation}
    DE_{Loss}= KL(\approxbetweencellpdffine, \betweencellpdffine) + \frac{1}{n} \cdot (\lambda_{\mu}\displaystyle\sum_{i}\mu_{raw}^2 + \lambda_{\sigma}\displaystyle\sum_{i}\sigma_{raw}^2)
\end{equation}
where $n$ is the number of coarse samples and $\lambda_{\mu}$ and $\lambda_{\sigma}$ are the regularization coefficients. We set the coefficient values to be in the range 0.01 to 0.1. The specific value depends on the number of samples along the rays (the specific value for $n$ samples is approximately $\frac{0.8}{n}$).\\

The $DE_{Loss}$ is added to the regular NeRF loss (\cref{nerf_loss}) so the overall loss is:
\begin{equation}
    L = L_{nerf} + \lambda_{DE}\cdot DE_{Loss}
\end{equation}
where $\lambda_{DE}$ is the $DE_{Loss}$ coefficient, set to be 0.1 in our experiments.

\subsection{Sampling and Smoothing}

\begin{figure}
\centering
\includegraphics[width=0.95\textwidth]{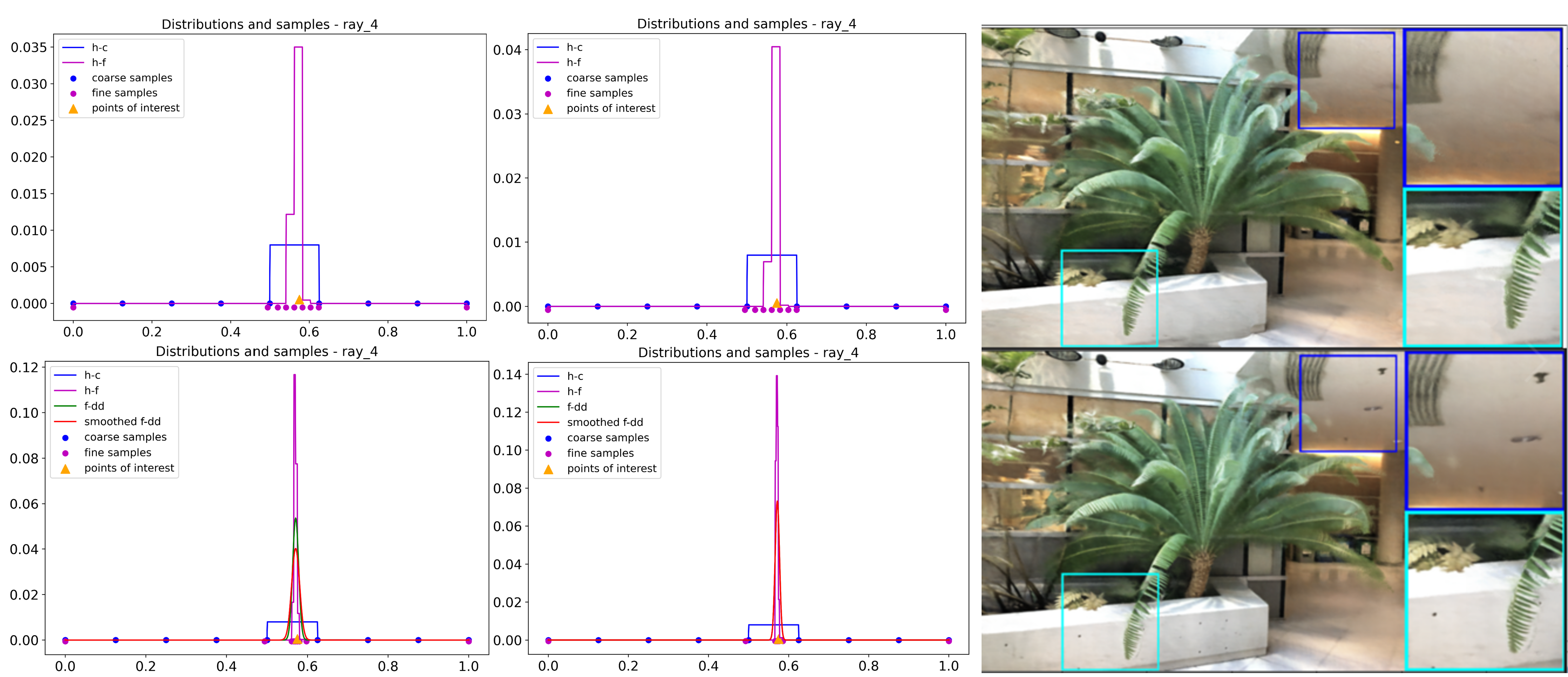}
\caption{\textbf{Density distribution during training}. Training with eight intervals: 50k iterations (left), 200k iterations (middle), RGB image after 200k iterations (right). The blue line is $\betweencellpdf$ and the green line is $\totalpdf$. The red line is the smoothed $\totalpdf$; note how the divergence between the red and green curves has closed during training as $u$ decreased toward 1. The blue dots are the coarse samples. The purple dots are the fine samples and the purple line is $\betweencellpdffine$. Our model (second row) keeps refining its fine sample's locations, while the MipNeRF sample's locations remain relatively similar from 50K to 200K iterations. Our model also achieves more accurate samples and a better RGB image than the regular model (first row). }
\label{fig:depth_dist}
\end{figure}

Except for the unbounded scene case, the first sampling stage in our model is always sampled uniformly along the ray. As in MipNerf, we use a 2-tap max filter followed by a 2-tap blur filter for smoothing $\betweencellpdf$ before sampling the second stage. For a small number of samples (up to 16), the smoothness method became a simple 1D blur filter with $[0.1,0.8,0.1]$ values during training, which helps us to achieve better accuracy in space (we found that this method works better for a small samples number in most scenes). For internal interval smoothing we defined an uncertainty factor $u \geq 1 $ that smooths the truncated $\incellpdfnormalized$ Gaussians inside the intervals by increasing $\sigma$: $\hat{\sigma} = u \cdot \sigma$. This uncertainty factor is decreased during training toward 1 and it corresponds to our increased certainty in the fine network depth estimation. Using this strategy also helps our model refine the second stage sample locations throughout the entire training process, while MipNeRF retains similar location from early stage in the training process. \Cref{fig:depth_dist} visualizes the differences between the two sampling methods. 

For an unbounded scene first pass, we also tried a different sampling strategy, of two methods. Similar to NeRF++\cite{kaizhang2020}, we dedicated half of the samples to uniformly sample the volume below radius 1 from the origin (the scene center). Outside the sphere we use the DONeRF log-sampling method \cite{neff2021donerf}. The second sampling stage remained the same. This method performed better for these scenes. Detailed results are described in Section 5.   

\section{Experiments}
We tested our model in three main domains: real-life forward facing scenes, synthetic $360\degree$ scenes and  real life $360\degree$ scenes. We compare the model's performance for different sampling budgets. We used the same number of samples for the coarse and fine networks. Thus, the number of samples listed in the results refers to one network. We used three different metrics when evaluating our results: structural similarity (SSIM), perceptual (LPIPS) and PSNR.\\
We divide our experiments into two parts. In part one, we focus on domains in which NeRF achieved excellent results: real-life forward facing and synthetic $360\degree$, part two contains domains that NeRF is struggling with: real life $360\degree$ bounded and unbounded. All our training used a single GeForce GTX 1080.

\subsubsection{Part 1:}
\begin{figure}
\centering
\includegraphics[width=0.85\textwidth]{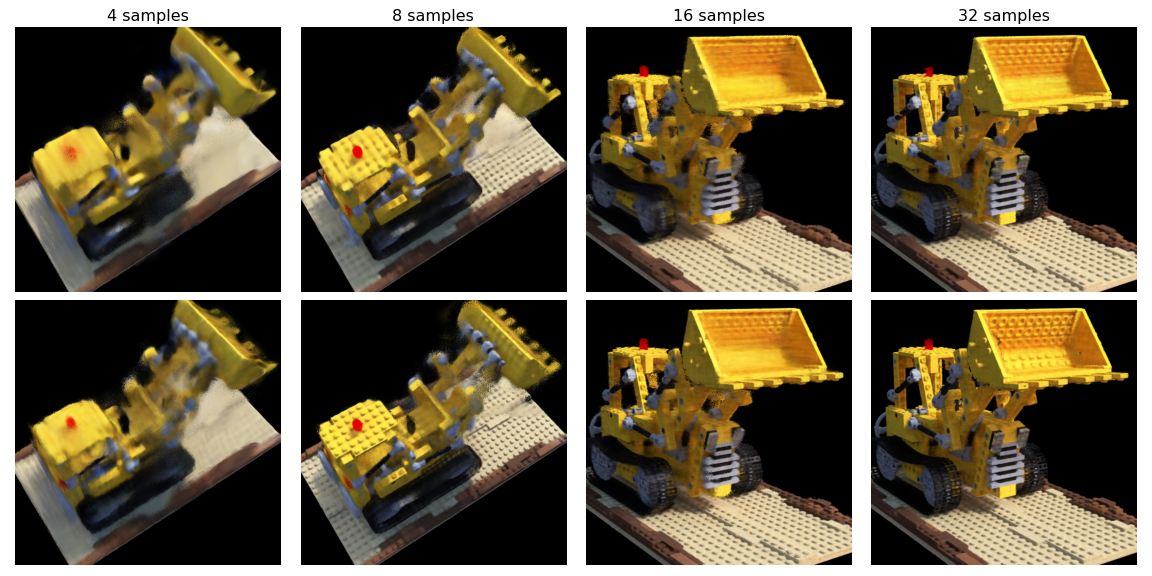}
\caption{\textbf{Lego results}: First row \textendash$~$MipNeRF model. Second row \textendash$~$our model. The number of samples is marked at the top of each column. Our model achieves better results for any number of samples.}
\label{fig:lego_results}
\end{figure}

For the forward facing scene we chose the fern scene from the LLFF paper \cite{mildenhall2019llff}. We used the NDC transformation as in NeRF and MipNeRF. For the synthetic $360\degree$ scene we chose the LEGO scene from the NeRF \cite{mildenhall2020nerf} example datasets. We trained each model with 200K iterations using 2048 rays per iteration. To challenge the model we reduced the number of samples and repeated the training routine several times. Each time we used different numbers of samples along the rays \textendash$~$4, 8, 16, 32. Validation was performed using the same number of samples as in the training. We compare our results with  those of MipNeRF, which trained and validated the model results under the same conditions.
Results are presented in \Cref{table:part_1_results} and in \Cref{fig:lego_results}.
Our model achieved better results in each one of the evaluation metrics for every number of samples.  
\begin{table}
\begin{center}
\caption{Experiment results on the LLFF fern dataset (real-world forward facing) and synthetic 360 deg LEGO scene. We trained each model for 200k iterations. Our model achieved better results than the regular models for every number of samples.}
\label{table:part_1_results}
\scalebox{0.9}{
\begin{tabular}{cccccccc}
\toprule

 &  & \multicolumn{3}{c}{FERN}     & \multicolumn{3}{c}{LEGO}  \\
                                        \cmidrule(lr){3-5}     \cmidrule(lr){6-8}
Samples & Model & PSNR$\uparrow$& SSIM$\uparrow$& LPIPS$\downarrow$  & PSNR$\uparrow$& SSIM$\uparrow$& LPIPS$\downarrow$ \\

\midrule
4  & MipNeRF & 20.20 & 0.521 & 0.606 & 21.64 & 0.733 & 0.281  \\ \cline{2-8} 
   & DDNeRF    & \textbf{20.81} & \textbf{0.577} & \textbf{0.507} & \textbf{21.79} & \textbf{0.741} & \textbf{0.274} \\ \hline
8 & MipNeRF  & 21.6 & 0.614 & 0.477 & 24.64 & 0.813 & 0.184 \\ \cline{2-8} 
   & DDNeRF    & \textbf{22.23} & \textbf{0.659} & \textbf{0.384} & \textbf{24.92} & \textbf{0.836} & \textbf{0.160} \\ \hline
16 & MipNeRF & 23.37 & 0.707 & 0.327 & 27.83 & 0.888 & 0.092 \\ \cline{2-8} 
   & DDNeRF    & \textbf{23.51} & \textbf{0.727} & \textbf{0.285} & \textbf{28.67} & \textbf{0.917} & \textbf{0.062} \\ \hline
32 & MipNeRF & 23.85 & 0.740 & 0.279 & 30.38 & 0.932 & 0.045 \\ \cline{2-8} 
   & DDNeRF    & \textbf{23.87} & \textbf{0.748} & \textbf{0.264} & \textbf{31.53} & \textbf{0.948} & \textbf{0.031} \\ 
\bottomrule
\end{tabular}}
\end{center}
\end{table}

\begin{figure}
\centering
\includegraphics[width=1.0\textwidth]{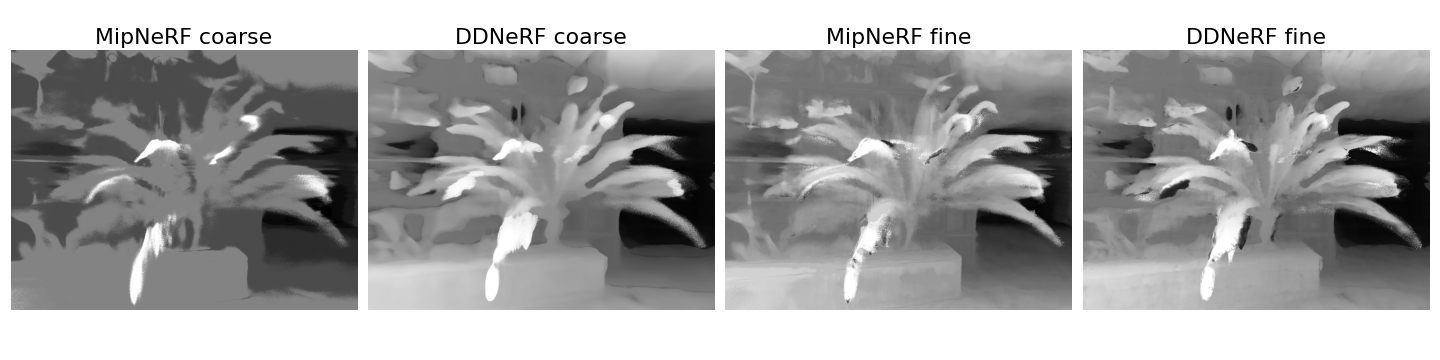}
\caption{\textbf{Disparity comparisons:} Comparison between MipNeRF ans DDNeRF estimated disparity on the fern scene. Both models were trained and evaluated using eight samples and without NDC warping. Notice how the depth estimation of our coarse model is closer to the fine model's estimation than to the MipNeRF coarse model.}
\label{fig:depth_comp_3}
\end{figure}

Another indication of the additional information our model gathers relative to the other models is its depth estimation. We extract the depth as the mean of the PDF along the ray, for the regular coarse model \textendash$~$ $\mathbb{E}[\betweencellpdf(t)]$ and for our coarse model \textendash$~$ $\mathbb{E}[\totalpdf(t)]$. For the fine models the depth calculated as $\mathbb{E}[\betweencellpdffine(t)]$.
Our coarse model produces a much better depth estimation than MipNeRF coarse model. \Cref{fig:depth_comp_3} shows the qualitative results.

\subsubsection{Part 2:}
\begin{figure}
\centering
\includegraphics[width=1\textwidth]{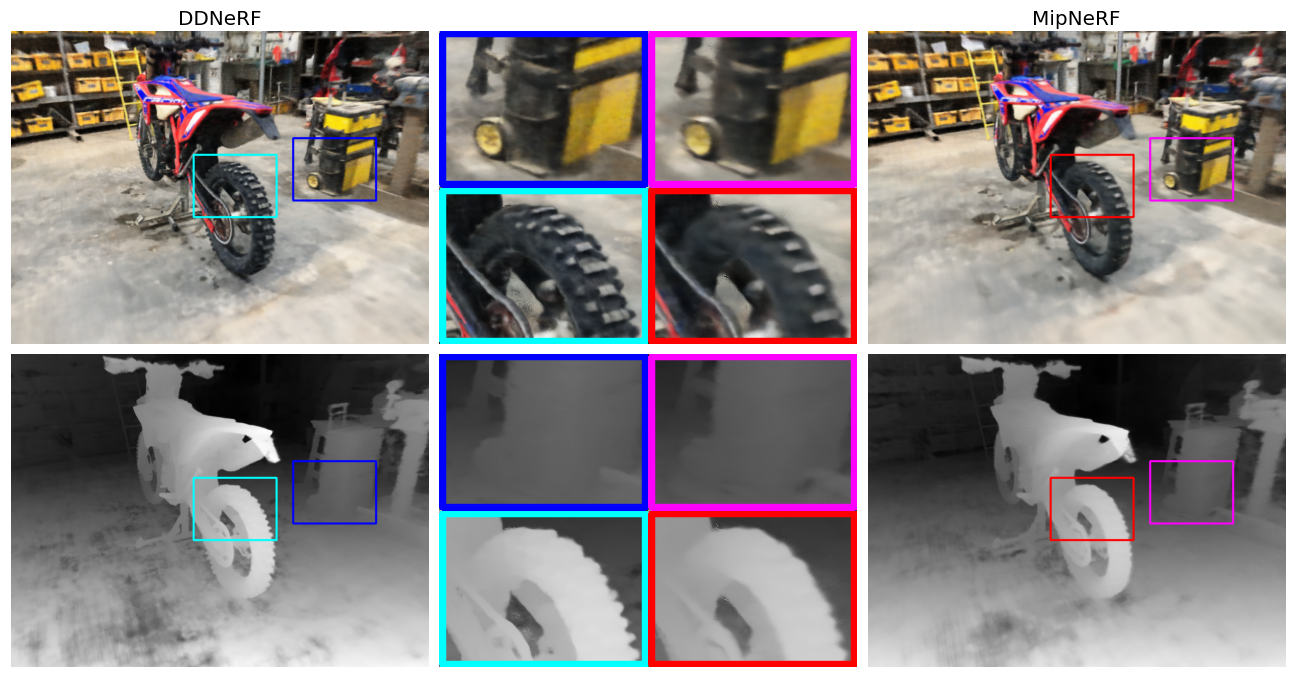}
\caption{\textbf{Motorcycle scene results:} First row\textendash$~$RGB predictions. Second row\textendash$~$disparity estimation from the fine model results. Our model achieves better depth estimation and better RGB prediction. Notice the crop for how our model is able to catch complex shapes such as the motorcycle's off road tires and the frame and small wheel of the tool cart.}
\label{fig:motorcycle_results}
\end{figure}

In the $360\degree$ domains we did not perform any space warping, excluding scale normalization of the world coordinate such that the main part of the scene is at a maximum distance of 1 from the origins.\\
For the bounded scene we created our own scene of a motorcycle inside a warehouse. We acquired 200 snapshots from $360\degree$ views, where $10\%$ of the images were saved for validation purposes. Although its depth is bounded, restoring this scene is not straightforward because it includes a big complex object and many small objects with fine details. We used the COLMAP structure from motion model \cite{schoenberger2016mvs} \cite{schoenberger2016sfm} to extract the relative orientation of the cameras. We trained each model with 300K iterations using 2048 rays per iteration. As in the first part of our experiment, we used a different number of samples. In this case \textendash$~$32, 64, 96. Results are shown in \Cref{table:part_2_results} and \Cref{fig:motorcycle_results}. As can seen from \Cref{table:part_2_results}, our model achieved better results in all metrics for any number of samples. More than that, our 32 sample model achieved better results than the 96 sample MipNeRF model. We can also see that our model produced more accurate depth estimation and better RGB prediction, especially around complex shapes (see \Cref{fig:motorcycle_results}).\\

\begin{table}
\begin{center}
\caption{Experiments results for real world $360\deg$ scenes. ``Smpl'' column stands for the number of samples in each of the networks (coarse and fine). All models were trained for 300K iterations. The left part of the table compares DDNeRF with MipNeRF for different numbers of samples. The right part compare alse to NeRF++ model.}
\label{table:part_2_results}
\scalebox{1}{
\begin{tabular}{ccccc|ccccc}
\toprule

 \multicolumn{5}{c}{Bounded scene \textendash$~$Motorcycle}     & \multicolumn{5}{c}{Unbounded scene \textendash$~$Playground}  \\
                                        \cmidrule(lr){1-5}     \cmidrule(lr){6-10}
Smpl & Model & PSNR$\uparrow$& SSIM$\uparrow$& LPIPS$\downarrow$ & Smpl & Model & PSNR$\uparrow$& SSIM$\uparrow$& LPIPS$\downarrow$ \\

\midrule
32 & MipNeRF & 20.36 & 0.533 & 0.532 & & MipNeRF & 21.47 & 0.547 & 0.540  \\ \cline{2-5} \cline{7-10} 
   & DDNeRF    & \textbf{20.84} & \textbf{0.577} & \textbf{0.453}  & 64 & DDNeRF & 21.71 & 0.568 & \textbf{0.498} \\ \cline{1-5}  \cline{7-10}
64 & MipNeRF  & 20.7 & 0.554 & 0.502  & & NeRF++ & \textbf{21.73} & \textbf{0.575} & 0.524 \\ \cline{2-10}
   & DDNeRF    & \textbf{21.07} & \textbf{0.592} & \textbf{0.422}  & & MipNeRF & 21.67 & 0.551 & 0.550 \\ \cline{1-5}  \cline{7-10}
96 & MipNeRF & 20.8 & 0.563 & 0.488  & 96 & DDNeRF & 21.69 & 0.569 & 0.498 \\ \cline{2-5} \cline{7-10}
   & DDNeRF    & \textbf{21.12} & \textbf{0.593} & \textbf{0.418}  & & NeRF++ & \textbf{21.74} & 0.589 &0.511 \\ \cline{2-5} \cline{7-10}
   &           &                &                &                 & & DDNeRF* & 21.43 & \textbf{0.596} & \textbf{0.451} \\
\bottomrule
\end{tabular}}
\end{center}
\end{table}

\begin{figure}
\centering
\includegraphics[width=0.9\textwidth]{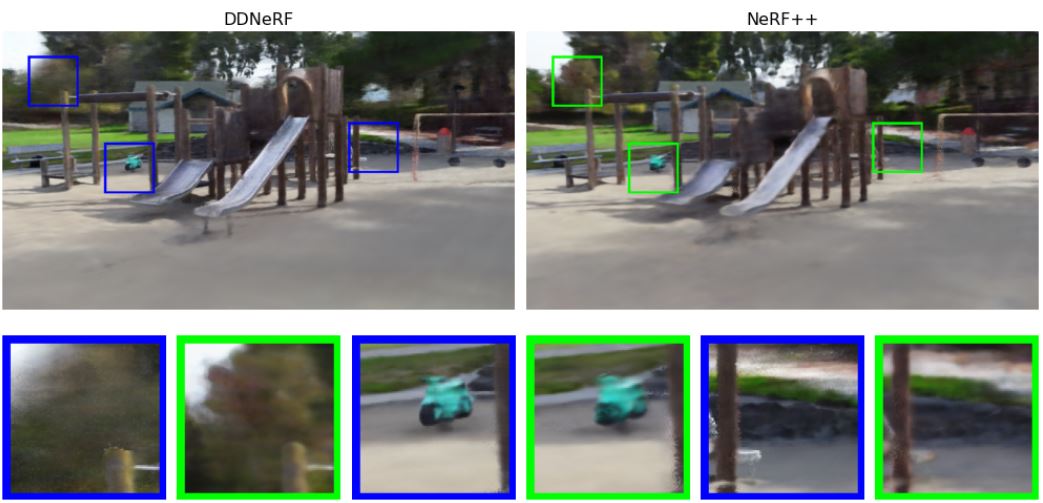}
\caption{DDNeRF vs NeRF++: Left image rendered using the DDNeRF* model, the right image using NeRF++. Notice that DDNeRF works better for the large foreground object. Two left crops: Far background \textendash$~$NeRF++ works better. Two middle crops: Small foreground objects \textendash$~$DDNeRF works better. Two right crops: Combination of foreground and close background  \textendash$~$DDNeRF works better.}
\label{fig:ddn_vs_npp}
\end{figure}

For an unbounded scene we chose the playground scene from the Tanks and Temples dataset \cite{Knapitsch2017}. We compare our model with both the MipNeRF and NeRF++ models. We trained and tested each model using 64 and 96 samples. For NeRF++, we split the sample budget equally between the foreground and the background models. For the 96 samples we also compared the unbounded scene sampling method that we described in Section 4.4. The model is notate as DDNeRF* in \Cref{table:part_2_results}. Our model achieved the best LPIPS and SSIM scores from the models we tested. NeRF++ achieved a better PSNR score (see \Cref{table:part_2_results}). When looking at the output images we can see that our model achieved better quality in the foreground and in the close background parts but struggles with far background parts; see \Cref{fig:ddn_vs_npp}

\setlength{\tabcolsep}{1.4pt}

\section{Conclusions}

In this paper we introduced DDNeRF. An extension of the MipNeRF model that produces more accurate representation of the density along the rays while improving the sampling procedure and the overall results. We showed that our model provides superior results on various domains and sample numbers. Our model uses fewer computational resources and produces better results.\\
Our new representation of the density distribution along the ray and our novel DE loss are general and can be adjusted to more NeRF variations. \\
For unbounded scenes, despite the good results we got, we believe that combining our model for foreground and close background together with the NeRF++ \cite{kaizhang2020} background model will lead to better results. We leave this for future work. \\

\textbf{Acknowledgements}
This work was supported by the Israeli Ministry of Science and Technology under The National Foundation for Applied
Science (MIA), and was partially supported by the Israel Science Foundation (grant No. 1574/21).

\clearpage
 
\bibliographystyle{splncs04}
\bibliography{main}

\end{document}